%% file: main_mobileSAMv2_arxiv.tex
\documentclass[10pt,letterpaper]{article}

\usepackage{cvpr}              

\input{preamble}

\definecolor{cvprblue}{rgb}{0.21,0.49,0.74}
\usepackage[pagebackref,breaklinks,colorlinks,citecolor=cvprblue]{hyperref}
\usepackage{multirow}
\usepackage{array}
\usepackage{graphicx}
\usepackage{amssymb}
\usepackage{wrapfig,lipsum,booktabs}
\makeatletter

\usepackage{hyperref}       
\usepackage{url}            

\newcommand{\thickhline}{%
    \noalign {\ifnum 0=`}\fi \hrule height 1pt
    \futurelet \reserved@a \@xhline
}
\newcolumntype{"}{@{\hskip\tabcolsep\vrule width 1pt\hskip\tabcolsep}}

\newcolumntype{C}[1]{>{\centering\arraybackslash}m{#1}}
\usepackage{colortbl}
\makeatother

\title{MobileSAMv2: Faster Segment Anything to Everything}

\author{
Chaoning Zhang\thanks{You are welcome to contact the authors through chaoningzhang1990@gmail.com} \\
	Kyung Hee University \\
\and
Dongshen Han \\
	Kyung Hee University\\
\and
Sheng Zheng \\
	Kyung Hee University\\
\and
Jinwoo Choi\\
	Kyung Hee University\\
\and
Tae-Ho Kim \\
	Nota Inc. \\
\and
Choong Seon Hong\\
Kyung Hee University\\
}

\begin{document}
\maketitle

\begin{abstract}
Segment anything model (SAM) addresses two practical yet challenging segmentation tasks: \textbf{segment anything (SegAny)}, which utilizes a certain point to predict the mask for a single object of interest, and \textbf{segment everything (SegEvery)}, which predicts the masks for all objects on the image. What makes SegAny slow for SAM is its heavyweight image encoder, which has been addressed by MobileSAM via decoupled knowledge distillation. The efficiency bottleneck of SegEvery with SAM, however, lies in its mask decoder because it needs to first generate numerous masks with redundant grid-search prompts and then perform filtering to obtain the final valid masks. We propose to improve its efficiency by directly generating the final masks with only valid prompts, which can be obtained through object discovery. Our proposed approach not only helps reduce the total time on the mask decoder by at least 16 times but also achieves superior performance. Specifically, our approach yields an average performance boost of 3.6\% (42.5\% \textit{v.s.} 38.9\%) for zero-shot object proposal on the LVIS dataset with the mask AR@$K$ metric. Qualitative results show that our approach generates fine-grained masks while avoiding over-segmenting things. This project targeting faster SegEvery than the original SAM is termed MobileSAMv2 to differentiate from MobileSAM which targets faster SegAny. Moreover, we demonstrate that our new prompt sampling is also compatible with the distilled image encoders in MobileSAM, contributing to a unified framework for efficient SegAny and  SegEvery. The code is available at the same link as MobileSAM Project \href{https://github.com/ChaoningZhang/MobileSAM}{\textcolor{red}{https://github.com/ChaoningZhang/MobileSAM}}.
\end{abstract}

\section{Introduction}
The NLP field has been revolutionalized by ChatGPT~\cite{zhang2023ChatGPT}, which constitutes a milestone in the development of generative AI (AIGC, a.k.a artificial intelligence generated content)~\cite{zhang2023complete}. GPT-series models~\cite{brown2020language,radford2018improving,radford2019language} trained on web-scale text datasets play a major role for its development. Following the success of foundation models~\cite{bommasani2021opportunities} in NLP, vision foundation models like CLIP~\cite{radford2021learning} have been developed by co-learning a text encoder via contrastive learning~\cite{he2020momentum,zhang2022dual}. More recently, a vision foundation model termed SAM~\cite{kirillov2023segment}, short for segment anything model, was released to solve two practical image segmentation tasks: segment anything (SegAny) and segment everything (SegEvery). Both two tasks perform class-agnostic mask segmentation, with the difference in what to segment. SegAny utilizes a certain prompt (like a point or box) to segment a single thing of interest in the image. By contrast, SegEvery aims to segment all things in the image. SAM has been widely used in a wide range of applications~\cite{zhang2023samsurvey} due to its impressive performance on these two tasks.

  \begin{figure}[!ht]
    \centering
     \begin{minipage}[b]{0.4\textwidth}
         \centering
         \includegraphics[width=\textwidth]{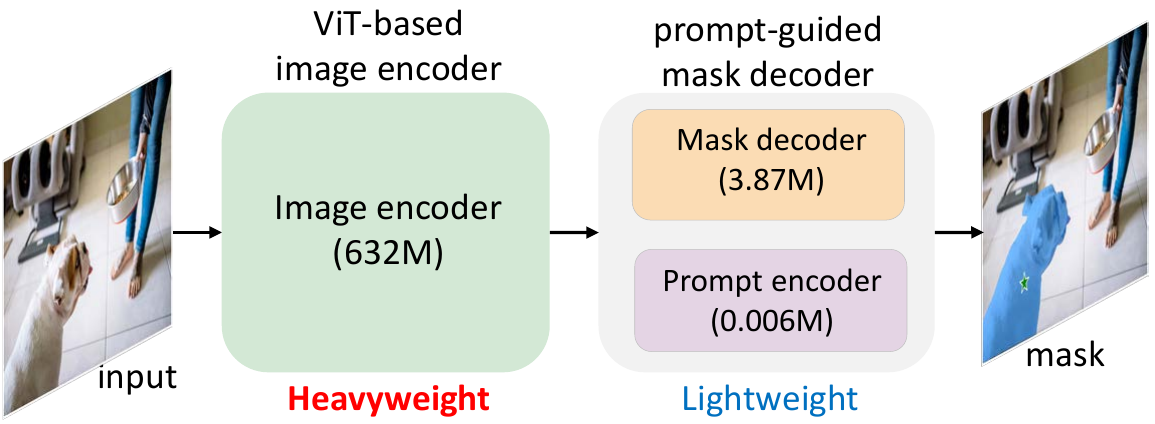}
     \end{minipage}
    \qquad
    \begin{minipage}[b]{0.49\textwidth}
    \begin{tabular}[b]{ccc}\hline
      Task & Image Encoder & Mask Decoder \\ \hline
      SegAny (1 point) &  $\sim 450$ms &  $\sim 4$ms  \\
      SegEvery($16 \times16$ points) & $\sim 450$ms &  $\sim 400$ms \\
      SegEvery($32 \times32$ points) & $\sim 450$ms &  $\sim 1600$ms \\
      SegEvery($64 \times64$ points) & $\sim 450$ms &  $\sim 6400$ms \\ \hline
    \end{tabular}
    \end{minipage}
    \captionlistentry[table]{A table beside a figure}
    \caption{SAM architecture and efficiency. The computation bottleneck for SegAny lies in its image encoder, while that for SegEvery mainly lies in its mask decoder when a high grid-search density is required (zero-shot object proposal in~\cite {kirillov2023segment} adopts $64 \times 64$ points).}
    \label{fig:architecture}
  \end{figure}

SAM works in sequence with two modules: ViT-based image encoder and prompt-guided mask decoder (see Figure~\ref{fig:architecture}). They are simply referred to image encoder and mask decoder in the remainder of this work when it does not confuse. The lightweight mask decoder adopts two-way attention to enable efficient interaction between image embedding and promt token for generating fine-grained masks~\cite {kirillov2023segment}. What makes SegAny slow is the image encoder which is 100+ more heavyweight than the mask decoder. This issue has been addressed by MobileSAM by distilling a lightweight image encoder in a decoupled manner. To segment all things, SegEvery requires first repeatedly running the mask decoder to generate numerous proposal masks and then selecting the high-quality and non-overlapping ones. This shifts the computation bottleneck from the image encoding to the mask generation and filtering. In essence, SegEvery is not a promptable segmentation task and thus the masks might be generated directly without using prompts~\cite{zhang2023faster}. Such a prompt-free approach has been attempted in~\cite{zhao2023fast}, which generates masks with less satisfactory boundaries (see analysis in Sec.~\ref{sec:prompt_free}). The mask decoder with two-way attention solves this problem but at the cost of making SegEvery much slower~\cite {kirillov2023segment}. To this end, we follow the practice of SegEvery in~\cite {kirillov2023segment} to prompt the mask decoder to guarantee the quality of the generated masks but address its low-speed issue by reducing the number of prompts.

SegEvery in~\cite {kirillov2023segment} prompts the image encoder with a grid search of foreground points. When the grid search is sparse, many small things or meaningful object parts might miss from being detected. Therefore, SegEvery in~\cite {kirillov2023segment} adopts a high grid density, like $64 \times 64$ points for zero-shot object proposal, which tends to have redundant prompts for large objects. In essence, it adopts a strategy to first generate many masks, most of which are redundant, and then filter the redundant ones. Intuitively, this process can be simplified by only generating valid masks, which saves time for mask generation and removes the need for mask filtering. Motivated by this intuition, we propose an efficient prompt sampling that seeks object-aware prompts. Fortunately, this is a well-solved issue in modern object detection. In this work, we adopt YOLOv8 which is a SOTA architecture for efficient detection with bounding boxes. To avoid over-fitting to any specific dataset, the model should be trained on an open-world dataset, for which a subset of SA-1B dataset is chosen. With the generated box, we can either use its center as an object-aware point prompt or directly adopt the box itself as the prompt. An issue with the point prompt is that it requires predicting three output masks per prompt to address the ambiguity issue. The bounding box is more informative with less ambiguity and thus is more suitable to be adopted in efficient SegEvery. Overall, this project is designed to make SegEvery in~\cite{kirillov2023segment} faster while achieving competitive performance. We term this project MobileSAMv2 to differentiate MobileSAM~\cite{zhang2023faster} that makes SegAny faster. Overall, the contributions of this work are summarized as follows.

\begin{itemize} 

\item We identify what makes SegEvery in SAM slow and propose object-aware box prompts to replace the default grid-search point prompts, which significantly increases its speed while achieving overall superior performance. 

\item We demonstrate that the our proposed object-ware prompt sampling strategy is compatible with the distilled image encoders in MobileSAM, which further contributes to a unified framework for efficient SegAny and SegEvery.

\end{itemize}

\section{Related Work} \label{sec:related}
\paragraph{Progress on SAM.} Since its advent in April 2023, SAM has been extensively studied in numerous GitHub projects and research articles. Its performance of SegAny, has been studied in various challenging setups, including medical images~\cite{ma2023segment,zhang2023input}, camouflaged objects~\cite{tang2023can}, and transparent objects~\cite{han2023segment}. Overall, SAM shows strong generalization performance but can be improved when the setup gets more challenging. Its generalization in the adversarial setup has been studied in Attack-SAM~\cite{zhang2023attacksam} which shows that the output masks of SAM can be easily manipulated by maliciously generated perturbations. Follow-up works further study the performance of adversarial perturbation generated on SAM in cross-model transferability~\cite{han2023segment} and cross-sample transferability~\cite{zheng2023black}. A comprehensive robustness evaluation of SAM has been studied in follow-up work~\cite{qiao2023robustness} which shows that SAM is robust against style transfer, common corruptions, local occlusion but not adversarial perturbation. The versatility of SAM has been demonstrated in another line of work. Even though SAM is shown to be compatible with text prompts in the original paper~\cite{kirillov2023segment} as a proof-of-concept, this functionality is not included in its official code. Grounded SAM~\cite{GroundedSegmentAnything2023} project combines Grounding DINO~\cite{liu2023grounding} with SAM for text-guided promptable segmentation. Specifically, Grounding DINO utilizes a box to generate a bounding box which can be used as a prompt for the SAM to predict a mask. Semantic segment anything project~\cite{chen2023semantic} introduces CLIP~\cite{radford2021learning} to assign labels to the predicted masks of SAM. SAM has also been shown to be versatile for image editing~\cite{rombach2022high}, inpainting tasks~\cite{yu2023inpaint} and object tracking in videos~\cite{yang2023track,z-x-yang_2023}. Beyond 2D, SAM can also be used for 3D object reconstruction~\cite{shen2023anything,kang2022any}, \textit{i.e.} assisting 3D model generation from a single image. PersoanlizeSAM~\cite{zhang2023personalize} personalizes the SAM with one shot for the customized SAM. High-quality tokens have been introduced in~\cite{ke2023segment} to improve the quality of predicted masks. \textit{The readers are suggested to refer to~\cite{zhang2023samsurvey} for a survey of SAM for its recent progress.} 

\paragraph{Class-agnostic segmentation.} Detection is a fundamental computer vision task that localize the objects of interest on an image~\cite{liu2020deep}. Detection roughly localizes the object by a box, while segmentation performs a more fine-grained localization by assigning a pixel-wise mask~\cite{minaee2021image}. It is straightforward to deduce a box from a given mask, but not vice versa, which indicates that the segmentation task is more complex than detection. Except for assigning masks, image segmentation (like semantic segmentation) often involves predicting their corresponding semantic labels from a predefined class set~\cite{guo2018review}. However, it is far from practical applications because there can be unlimited classes in the real world. To this end, a line of work has attempted to extend them to the open world by not considering their semantic labels. Class-agnostic object detection has been first formally proposed in~\cite{jaiswal2021class} with the average recall established as the metric to evaluate its performance and then be used as a new pretraining technique~\cite{bar2022detreg}. Multimodal transformer has been shown in~\cite{maaz2022class} to demonstrate satisfactory performance. Open-world instance segmentation has been extensively in~\cite{kim2022learning,wang2022open,wu2023exploring} for realizing class-agnostic detection and segmentation. In contrast to them treating the object as a whole, a follow-up work~\cite{pan2023towards} has investigated open-world object part segmentation. More recently, SAM~\cite{kirillov2023segment} has solved the SegEvery task that segments all things including all objects and their meaningful parts. It has been shown in multiple Github projects (CLIP-SAM, Segment-Anything-CLIP, segment-anything-with-clip) that class-agnostic segmentation masks obtained from SegEvery with SAM~\cite{kirillov2023segment} can be combined with CLIP~\cite{radford2021learning} to produce semantic-aware segmentation in the open world.

\section{Segment Everything} \label{sec:background}

\paragraph{Task Definition.} Conventional image segmentation predicts pixel-wise masks together with their corresponding class labels. However, the classes can be ambiguous across different datasets. For example, CIFAR10 dataset has a \textit{dog} class, while ImageNet-1K has several hundred classes to indicate various breeds of dogs. Another setup might divide them into puppy or adult dogs instead of their breed. This makes open-world image segmentation not tractable when considering the semantics. When decoupled from label prediction, open-world image segmentation becomes relatively easier but remains a challenging issue. Without semantic information, whether a region in the image is considered an object or a thing denoted by a mask can be subjective. This ill-posed nature is, at least partly, connected to the ambiguity of granularity~\cite{li2023semantic}. For example, when the granularity is too large, it might only detect a large object but ignore its meaningful object parts. When the granularity is too small, every pixel can be independently segmented, which is trivial and meaningless. In other words, open-world image segmentation requires segmenting all things including the whole objects and their meaningful parts, \textit{i.e. \textbf{everything}}. In essence, it is a class-agnostic segmentation task that performs \textit{zero-shot object proposal generation} in the \textit{open world.}  This task is termed \textbf{\textit{segment everything}} (SegEvery) in~\cite{kirillov2023segment}, and we follow~\cite{kirillov2023segment} to adopt the same name to avoid confusion.

\paragraph{Prompt-aware Solution.} SAM is a pioneering work to solve the task of promptable segmentation~\cite{kirillov2023segment}. Specifically, it segments any object of interest with a certain prompt, which is named segment anything (SegAny) in~\cite{kirillov2023segment}. Based on this, SAM provides a straightforward solution to the SegEvery task by prompting the SAM decoder with a search grid of foreground points. An underlying issue of this approach is that the performance is highly dependent on the grid density. Intuitively, a higher grid density tends to yield higher performance but at a cost of significantly increasing the computation overhead. Orthogonal to MobileSAM~\cite{zhang2023faster} distilling the heavyweight image encoder for faster SegAny, this project, named MobileSAMv2 for term differentiation, aims to make SegEvery faster by proposing a new sampling strategy to reduce the number of sampled prompts. Our solution significantly improves its efficiency while achieving overall superior performance. In the following section, we will illustrate the motivation behind our solution and its detailed implementation.

\section{Method}
\subsection{Motivation and Framework}
The prompt-aware solution proposed in~\cite{kirillov2023segment} has demonstrated impressive performance for the challenging SegEvery task. It adopts a strategy of first generating redundant masks and then filtering them to obtain the final valid masks. Intuitively, this process might be unnecessarily cumbersome and can be simplified by prompting the mask decoder with only valid prompts, which saves time for mask generation and has no need to perform any filtering. The core of our method lies in replacing the default gird-search prompt sampling with object-aware prompt sampling. This strategy boils down to determining whether there is an object in a certain region on the image. Modern object detection task already solves this by localizing the objects with bounding boxes. Most of the generated bounding boxes overlap with each other, which thus requires pre-filtering before being used as valid prompts. Without additional prior knowledge, we deduce the filter-left bounding box center as the foreground point with a moderate assumption that the box center point is on the object. Moreover, the mask decoder of SAM also accepts a box as the prompt. Therefore, we also experiment with directly using the remaining box as the prompt. Overall, our proposed SegEvery framework consists of two stages: object-aware prompt sampling and prompt-guided mask decoding. The first stage samples the prompts by relying on a modern object detection network, and the second stage follows SAM~\cite{kirillov2023segment} to perform a prompt-guided mask decoding.

\subsection{Object-Aware Prompt Sampling}

Object discovery has been widely used in some cases (like visual-language tasks) as a preprocessing technique for avoiding exhaustive sliding window search. Inspired by their practice, we propose to exploit object discovery for sampling prompts. In essence, object discovery is to localize the objects with a bounding box, which can be realized by modern object detection models but excluding its classification head. The past decade has witnessed a huge advancement in the development of object detection models, YOLO family models have become de facto standard choice for its advantages in real-time performance. To prevent over-fitting to any specific domain, the chosen YOLOv8 model needs to be trained on an open-world dataset, for which a small subset of SA-1B dataset~\cite{kirillov2023segment,zhang2023faster} is chosen. The model is trained with the supervision of both the bounding box and masks and then finetuned with only the bounding box loss. Such a training approach also facilitates comparison with the prompt-free approach (see Sec.~\ref{sec:prompt_free}). This generates numerous overlapping boxes, which need to be filtered before being used as prompts. Following the standard practice, we adopt NMS to filter the overlapping boxes. With the filtered bounding boxes, we can either use its center as an object-aware point prompt or directly adopt the box itself as the prompt. In practice, we choose the latter strategy for multiple reasons. Even though the center point is object-aware, it is based on an assumption that the object inside the bounding box covers the center point. This holds in most cases but not in all cases. Another issue with the point prompt is that it needs to predict three output masks to address the ambiguity issue, which requires additional mask filtering. By contrast, the box prompt is more informative and generates high-quality masks with less ambiguity, which mitigates the need to predict three masks and is thus more beneficial for efficient SegEvery.

\subsection{Prompt-guided Mask Decoding} 

We follow SAM~\cite{kirillov2023segment} to perform a prompt-guided mask decoding in a batch manner. In contrast to the image encoder setting the number of image samples as batch, here, the batch concept is the number of prompts. It is worth noting that the prompt-guided mask decoder in SAM also accepts a box as the input. Therefore, it is technically feasible to directly prompt the mask decoder with a set of boxes that save the process of deriving the center points. Even though it is not our original motivation, without causing any additional cost, we find that this practice yields a non-trivial performance boost. In other words, it can be seen as a free trick to improve the task performance. Prompt-aware solution in~\cite{kirillov2023segment} requires mask filtering. Empirically, we find that this process can be very slow because the mask is high-dimensional. This is different from efficient box filtering because a box only has four dimensions. This cumbersome mask filtering is optional in our proposed SegEvery framework because we can avoid it by prompting the mask decoder with only valid prompts. In other words, we keep all the generated masks since the prompts are sampled in an object-aware manner.

\section{Experiments}

SegEvery has been perceived in~\cite{kirillov2023segment} as a zero-shot object proposal task with standard average recall (AR) as the metric for performance evaluation. We follow the practice in~\cite{kirillov2023segment} to adopt AR for masks at $K$ proposals (mask AR@$K$), where $K$ is the maximum allowable number of masks. With the definition of AR, AR@$K$ gets higher when $K$ is allowed to set to a larger value, which constitutes a less strict metric. Only AR@$1000$ is reported in~\cite{kirillov2023segment}, but we choose to report AR@$K$ for $K$ ranging from $10$ to $1000$. To not lose generality yet save computation resources, we choose to report the results on 100 images randomly sampled from the large vocabulary instance segmentaiton (LVIS) dataset~\cite{gupta2019lvis}.

\subsection{Main Results} 

What makes SegEvery much more computation-intensive than SegAny lies in the need to run the mask decoder with numerous sampled prompts~\cite{kirillov2023segment}. Our proposed object-aware prompt sampling improves its efficiency by reducing the number of total prompts. In the following, we detail their difference in terms of required computation time by roughly dividing the prompt-guided mask decoding pipeline into two stages: prompt encoding (including pre-sampling) and mask decoding (including post-filtering). Mask decoding is much more heavy than simple prompt encoding. Except for the redundant sampled prompts, SegEvery in~\cite{kirillov2023segment} generates more masks than needed (or AR@$K$ allowed) by setting the multi-mask option to true. Specifically, one point can prompt the mask decoder to generate three output masks with different granularities (small, middle, and large). Setting the multi-mask option to true has been demonstrated in~\cite{kirillov2023segment} for achieving superior performance of SegEvery (like zero-shot object proposal) but at the cost of requiring filtering more redundant masks. Empirically, we find that (GPU-processed) mask filtering in~\cite{kirillov2023segment} can be even more computation insensitive than the mask generation itself partly because accessing and removing high-dimensional masks on the memory can be very slow. Interested readers are suggested to check their official code for details. Intuitively, the time spent on the mask decoder for the grid-search prompt sampling in~\cite{kirillov2023segment} depends on the grid density (See Figure~\ref{fig:architecture}). Different tasks might require different grid densities. In the official demo~\cite{kirillov2023segment}, it adopts a grid density of $32 \times 32$ which achieves a good trade-off between efficiency and performance. For evaluating the performance on zero-shot object proposal, a grid density of $64 \times 64$ is adopted in~\cite{kirillov2023segment}.

\textbf{Efficiency comparison.} SegEvery with our proposed sampling strategy needs to run an object discovery algorithm to obtain object-aware prompts, which requires more time for prompt sampling than the default grid-search sampling in~\cite{kirillov2023segment} but needs to encode much fewer prompts. For the mask generation, the time spent on the mask decoder is somewhat proportional to the number of sampled prompts. We find that the performance saturates when the number of prompts is approaching 320, which is set to the maximum number of detection boxes (See Sec.\ref{sec:ablation}). Less computation is needed when the object discovery generates masks that are fewer than 320, which occurs in many cases. Nonetheless, when performing an efficiency analysis, we compare our most computation-intensive scenario (max 320 prompts) with the grid-search strategy. The results in Table~\ref{tab:main_results_time} show that our proposed prompt sampling strategy significantly improves the efficiency of the (prompt-guided) mask decoder by at least 16 times. The computation spent on the prompt encoding accounts for a non-trivial percentage with our approach because it runs object discovery for prompt sampling. A more efficient object discovery algorithm is left for future work.

\begin{table}[!htbp]
\centering
\caption{Efficiency comparison of the (prompt-guided) mask decoder between grid-search sampling and object-aware sampling. Note that the prompt encoding includes the prompt pre-sampling time, while the mask decoding includes the mask post-filtering time.}
\label{tab:main_results_time}
\scalebox{0.8}{
\begin{tabular}{ccccccc}
\toprule
Sampling strategy & Prompt Encoding & Mask Decoding & Total \\ 
\midrule
Grid-search sampling ($32 \times 32$ prompts) & 16ms & 1600ms & 1616ms \\
Grid-search sampling ($64 \times 64$ prompts) & 64ms & 6400ms & 6464ms \\
\midrule
Object-aware sampling (max $320$ prompts) & $47$ms & 50ms & 97ms \\
\bottomrule
\end{tabular}
}
\end{table}

\textbf{Performance comparison.} We carefully follow the implementation practice recommended in~\cite{kirillov2023segment} for zero-shot object proposal. By default, it is suggested to set the grid density to $64\times 64$ and generate a total of $12288$ ($64\times 64\times 3$) masks, out of which a maximum of 1000 masks are then selected given the mask AR@$1000$ metric. We have experimented with decreasing the grid density and/or setting the multi-mask option to false (single-mask mode). The results in Table~\ref{tab:main_results_mask1000} show that generating fewer masks by either one of the above two practices leads to a performance drop, suggesting that the default grid-search sampling strategy highly relies on generating redundant masks for selecting the final needed ones. Moreover, we have multiple major observations by comparing SAM (the default grid-search prompt sampling) and MobileSAMv2 (our proposed object-aware prompt sampling). First, under the condition of prompting with the same type of prompt (points) and setting multi-mask to false, we find that MobileSAMv2 (max 320 points) achieves comparable performance as SAM using 4096 points, suggesting that the object-aware property of our prompt sampling strategy significantly avoids redundancy. Boosted with the multitask option set to true, the default $64 \times 64$ grid density yields a higher performance (59.2\%), which constitutes the best setup for the grid-search strategy. Similarly, we can also increase the performance of our object-aware point sampling by setting the multi-mask to true. Note that the motivation for predicting three output masks of different granularities~\cite{kirillov2023segment} is to address the ambiguity issue of a point prompt. A single point has limited prompt information and thus causing ambiguity (the readers can check Figure 4 in~\cite{kirillov2023segment} for more details). By contrast, a box prompt is much more informative and reduces ambiguity to a very large extent. This is supported by our results in Table~\ref{tab:main_results_mask1000} that box prompts yield a significant performance boost at single mask mode. Last, it is worth mentioning that, compared with the best result of the grid-search sampling strategy (with $64 \times 64$ points at multi-mask mode), our proposed sampling strategy (with max 320 box prompts) achieves comparable performance (59.3\% \textit{v.s.} 59.2\%). Limiting the max number of prompts to 256, our strategy still yields competitive performance (58.5\%) compared with that of the grid-search strategy (34.6\%) under the same condition. We also report AR@$K$ for other $K$ values in Table~\ref{tab:main_results_maskk}. When $K$ is set to a relatively small value, we find that our proposed object-aware sampling strategy with much fewer prompts leads to a performance boost by a large margin. Overall, our proposed approach achieves an average performance boost of 3.6\% (42.5\% \textit{v.s.} 38.9\%).

\begin{table}[!htp]
\centering
\caption{Zero-shot object proposal comparison between grid-search sampling and object-aware sampling (mask@1000 as the metric).}
\label{tab:main_results_mask1000}
\scalebox{0.8}{
\begin{tabular}{ccccccc}
\toprule
Method  &  multi-mask  &all & small & med. & large\\ 
\midrule
SAM($64 \times 64 = 4096$ points) &  true ($\times 3$) & 59.2 & 46.6 & 78.7 & 82.4 \\
SAM($32 \times 32 = 1024$ points) &  true ($\times 3$) & 57.2 & 42.9 & 79.2 & 83.6\\ 
SAM($16 \times 16 = 256$ points) &  true ($\times 3$) & 40.0 & 19.4 & 71.3 & 79.1\\ 
SAM($64 \times 64 = 4096$ points) &  false ($\times 1$)  & 54.3 & 44.4 & 71.5 & 67.4\\
SAM($32 \times 32 = 1024$ points) &  false ($\times 1$) & 49.8 & 37.2 & 71.4 & 66.8\\ 
SAM($16 \times 16 = 256$ points) &  false ($\times 1$) & 34.6 & 17.5 & 61.5& 64.9\\ 
\midrule
MobileSAMv2 (max $320$ points) &  true ($\times 3$)   & 55.7 & 40.6& 78.6 & 84.6 \\
MobileSAMv2 (max $320$ points) &  false ($\times 1$) & 53.6 & 44.0& 70.4 & 66.6 \\
MobileSAMv2 (max $320$ boxes) &  false ($\times 1$)   &59.3 &47.9 &77.1 & 79.9 \\ 
MobileSAMv2 (max $256$ boxes) &  false ($\times 1$)   &58.5 &46.7 &77.1 & 79.1\\ 

\bottomrule
\end{tabular}
}
\end{table}

\begin{table}[!htp]
\centering
\caption{Zero-shot object proposal comparison between grid-search sampling and object-aware sampling.}
\label{tab:main_results_maskk}
\scalebox{0.8}{
\begin{tabular}{ccccccc}
\toprule
& Method  &  multi-mask  &all & small & med. & large\\ 
\midrule
mask AR@$1000$& SAM($64 \times 64 = 4096$ points) &  true ($\times 3$) & 59.2 & 46.6 & 78.7 & 82.4 \\
mask AR@$1000$& MobileSAMv2 (max $320$ boxes) &  false ($\times 1$)   &59.3 &47.9 &77.1 & 79.9\\ 
\midrule
mask AR@$100$& SAM($64 \times 64 = 4096$ points) &  true ($\times 3$) & 44.8 & 29.8 & 67.6 & 73.8 \\
mask AR@$100$& MobileSAMv2 (max $100$ boxes)  &  false ($\times 1$)  &50.6 &36.3&73.1 & 76.3\\ 
\midrule
mask AR@$10$& SAM($64 \times 64 = 4096$ points) &  true ($\times 3$) & 12.6 & 2.9 & 22.7 & 45.0 \\
mask AR@$10$& MobileSAMv2 (max $10$ boxes)  &  false ($\times 1$)  &17.6 &6.4 &35.0 & 37.8\\ 
\midrule
average& SAM($64 \times 64 = 4096$ points) &  true ($\times 3$) & 38.9 & 26.43 & 56.3 & 67.1 \\
average& MobileSAMv2   &  false ($\times 1$)  &42.5 &30.2 &61.7 & 64.7\\ 
\bottomrule
\end{tabular}
}
\end{table}

\begin{table}[!htp]
\centering
\caption{Influence of the image encoders on MobileSAMv2 for zero-shot object proposal (mask@1000).}
\label{tab:object_proposal_encoder}
\scalebox{0.8}{
\begin{tabular}{ccccccc}
\toprule
Encoder  & All & small & med. & large\\ 
\midrule
ViT-H &59.3 &47.9 &77.1 & 79.9 \\
TinyViT  &51.1 &38.9 &69.9 &73.4 \\ 
EfficientViT-L2 & 56.3 & 44.7 & 74.1 & 78.1\\ 
\bottomrule
\end{tabular}
}
\end{table}

 \subsection{On the Compatibility with Distilled Image Encoders} \label{sec:encoder}
In the above, we only consider the prompt-guided mask decoder, however, the whole pipeline needs to run the image encoder once before running the mask decoder. As shown in Figure~\ref{fig:architecture}, the time spent on the image encoder is relatively small for SegEvery with the grid-search point sampling. However, this is no longer the case when adopting our object-aware prompt sampling strategy, which reduces the time on the mask decoder to around 100ms. Therefore, we consider reducing the time spent on the image encoder by replacing the original one (ViT-H) in the SAM with a distilled one in the MobileSAM project~\cite{zhang2023faster}. The results with different distilled image encoders are shown in Table~\ref{tab:object_proposal_encoder}. We observe a moderate performance drop (from 59.2\% to 56.3\%) when EfficientViT-L2 is used. Given that EfficientViT-l2 runs around 20ms which is significantly faster than that of ViT-H (more than 400ms), it is worthwhile to replace the image encoder. Due to the simplicity and effectiveness of decoupled knowledge distillation introduced in MobileSAM~\cite{zhang2023faster}, a more powerful distilled image encoder is expected to emerge soon to further alleviate the performance drop. It is worth highlighting that MobileSAM and MobileSAMv2 solve two orthogonal issues: faster SegAny and faster SegEvery. Combing them together constitutes a \textit{unified} framework for efficient SegAny and SegEvery.

\section{Additional Comparison and Ablation Study}

\subsection{Comparison with Prompt-free Approach} \label{sec:prompt_free}

\begin{table}[!htp]
\centering
\caption{Zero-shot object proposal comparison between prompt-free and prompt-aware approaches (mask@1000).} 
\label{tab:comparison_prompt_free}
\scalebox{0.8}{
\begin{tabular}{ccccccc}
\toprule
Strategy & Method  & All & small & med. & large \\ 
\midrule
Prompt-free & FastSAM  & 49.6 &36.2 & 69.4 & 77.1  \\ 
\midrule
Prompt-aware & SAM(best setup) & 59.2 & 46.4 & 78.7 & 82.4 \\
Prompt-aware & MobileSAMv2 (ViT-H) &59.3 &47.9 &77.1 & 79.9 \\
Prompt-aware & MobileSAMv2 (EfficientViT-L2 ) & 56.3 & 44.7 & 74.1 & 78.1\\ 

\bottomrule
\end{tabular}
}
\end{table}

\begin{figure*}[!htbp]
    \centering
     \begin{minipage}[t]{0.23\textwidth}
         \includegraphics[width=\textwidth]{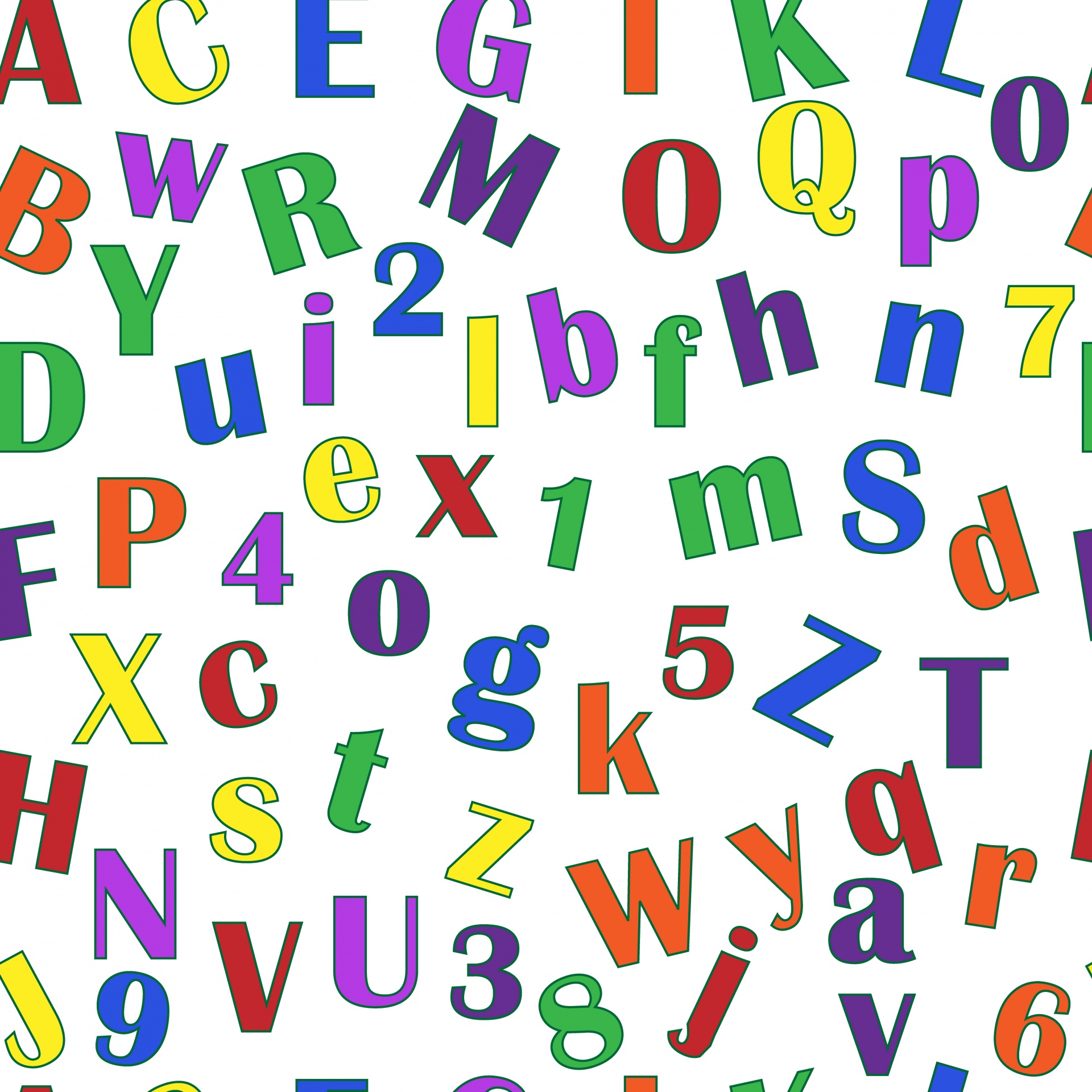}
     \end{minipage}
    \begin{minipage}[t]{0.23\textwidth}
         \includegraphics[width=\textwidth]{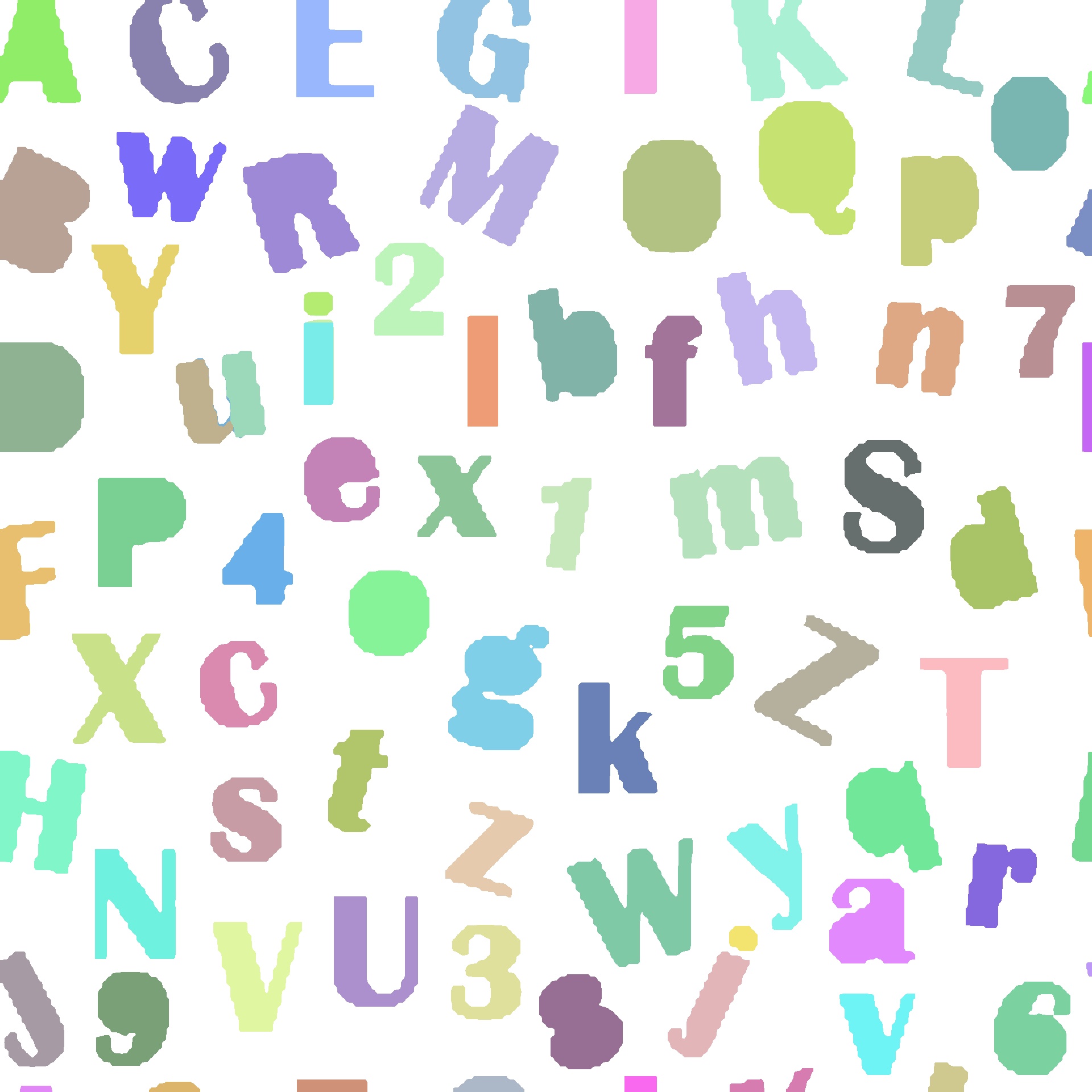}
     \end{minipage}
    \begin{minipage}[t]{0.23\textwidth}
         \includegraphics[width=\textwidth]{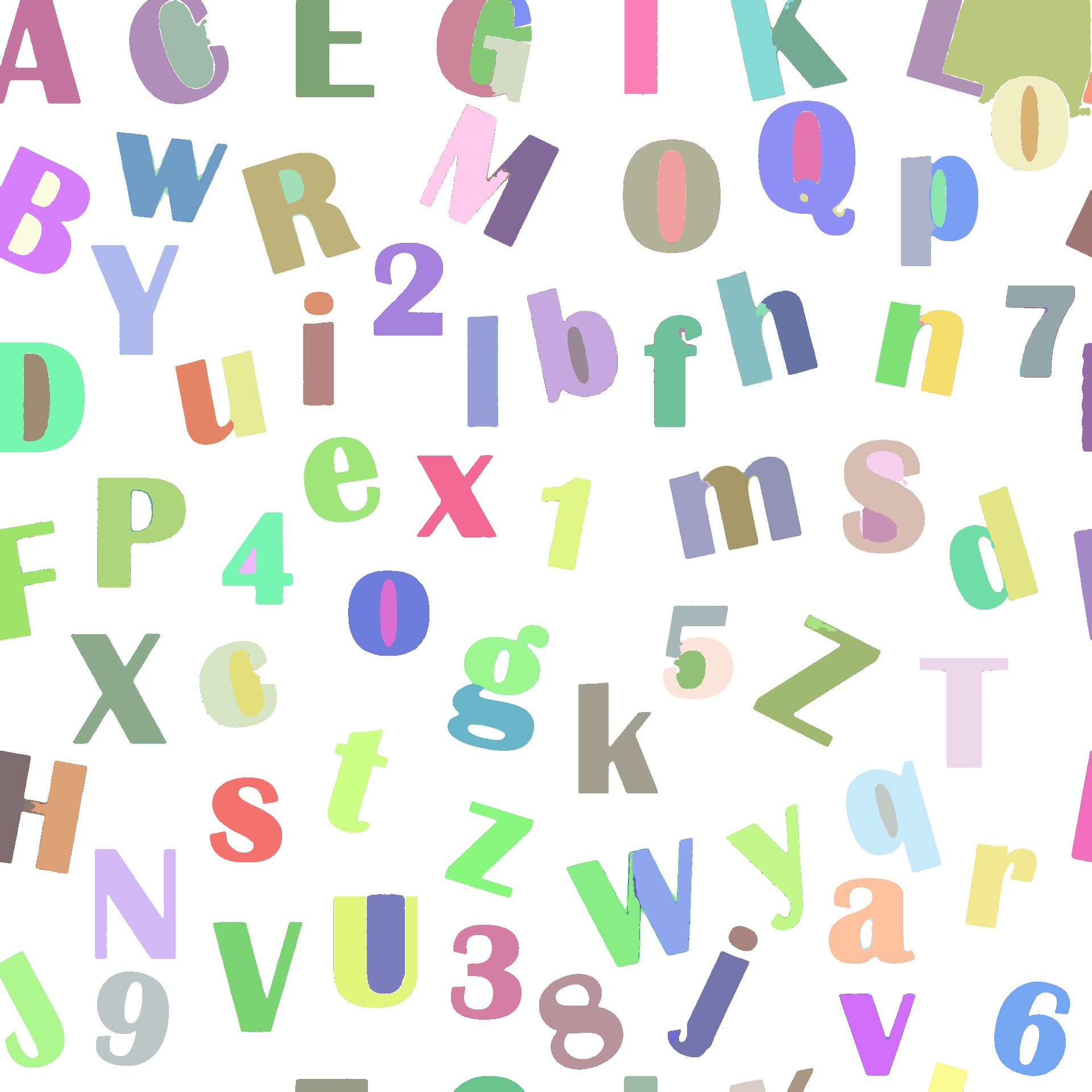}
     \end{minipage}
    \begin{minipage}[t]{0.23\textwidth}
         \includegraphics[width=\textwidth]{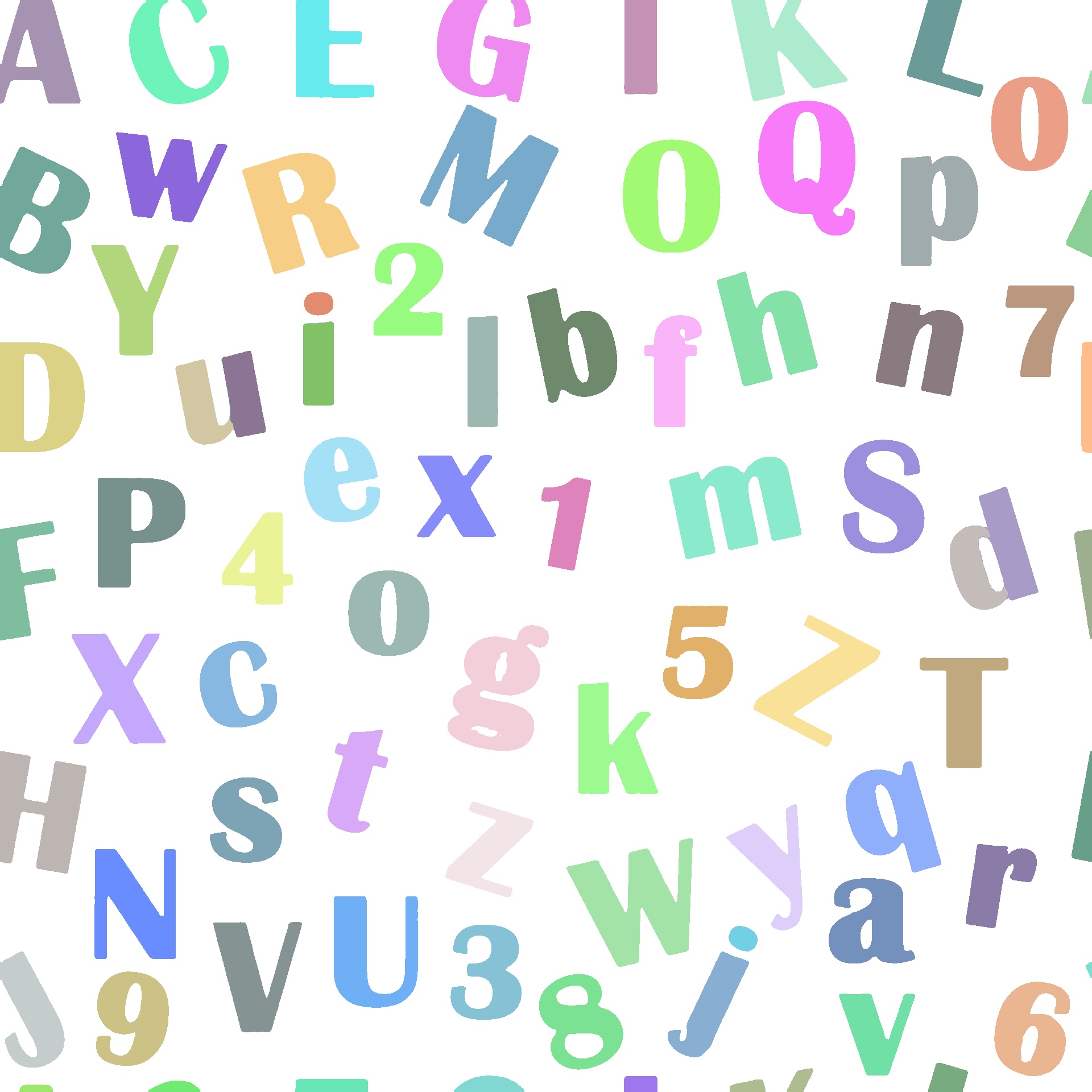}
     \end{minipage}

    \begin{minipage}[t]{0.23\textwidth}
         \includegraphics[width=\textwidth]{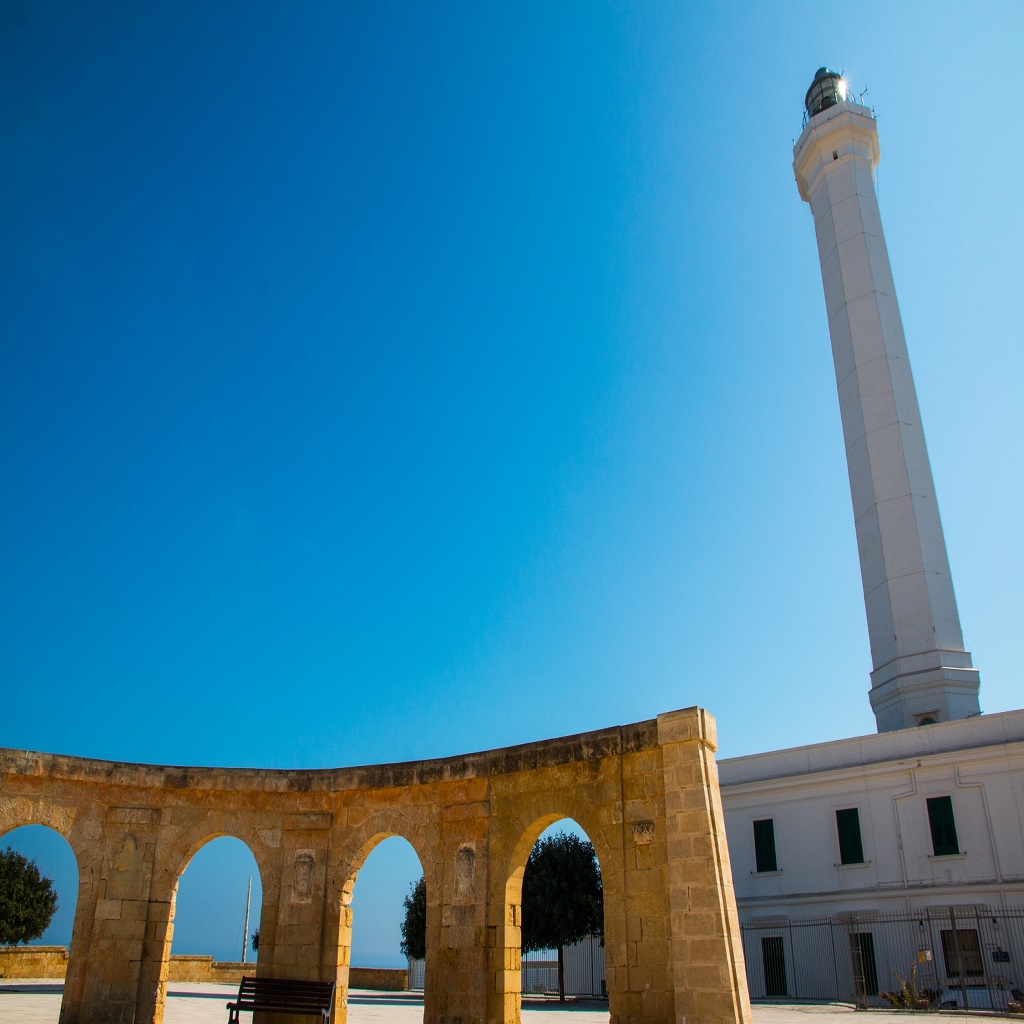}
         \subcaption{Original Image}
     \end{minipage}
    \begin{minipage}[t]{0.23\textwidth}
         \includegraphics[width=\textwidth]{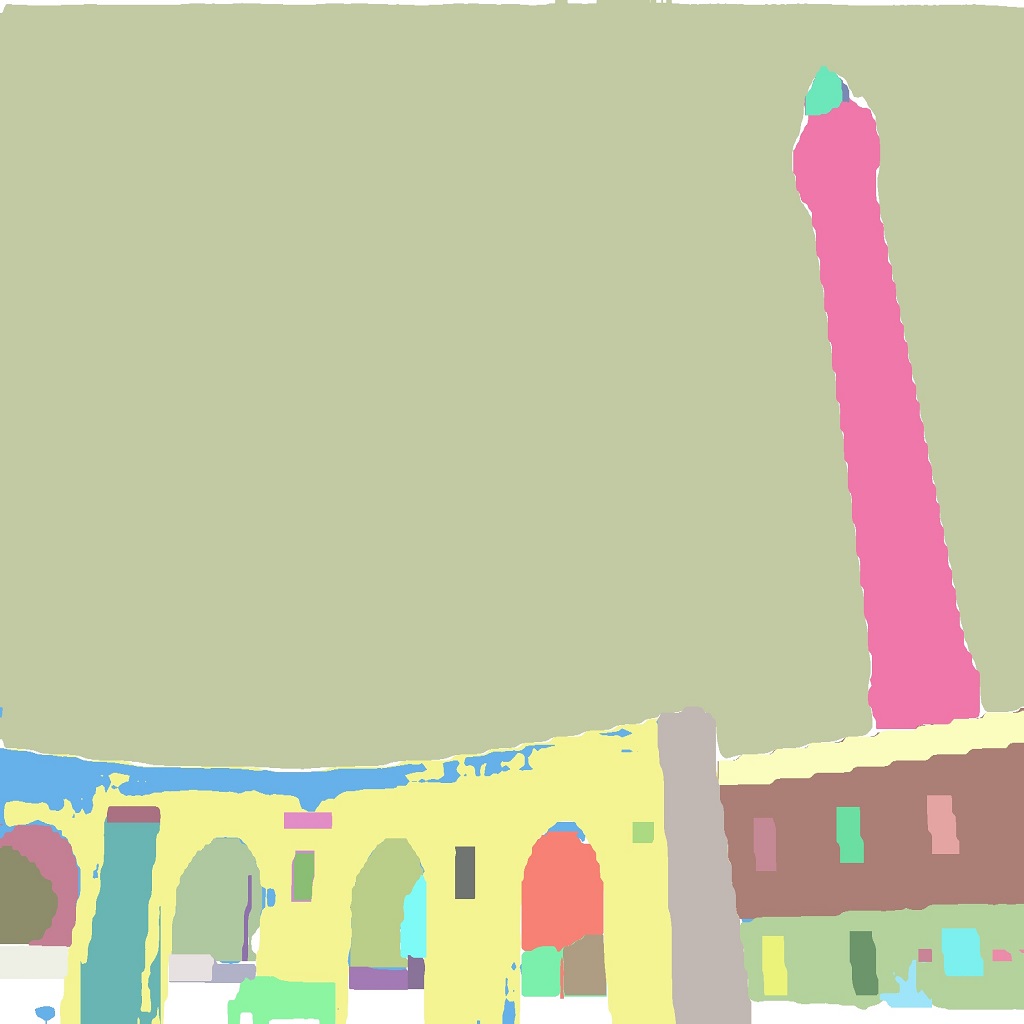}
         \subcaption{Prompt-free (FastSAM)}
     \end{minipage}
    \begin{minipage}[t]{0.23\textwidth}
         \includegraphics[width=\textwidth]{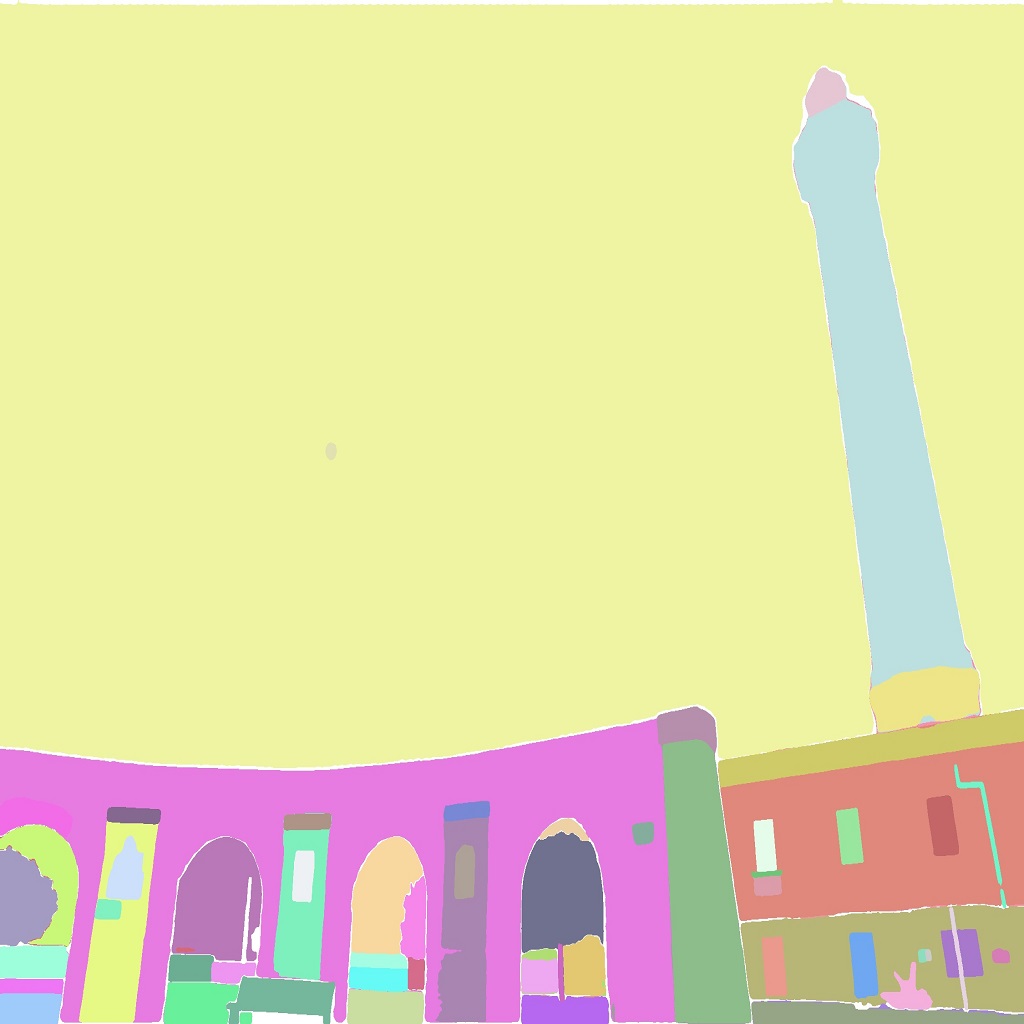}
         \subcaption{Prompt-aware (SAM))}
     \end{minipage}
    \begin{minipage}[t]{0.23\textwidth}
         \includegraphics[width=\textwidth]{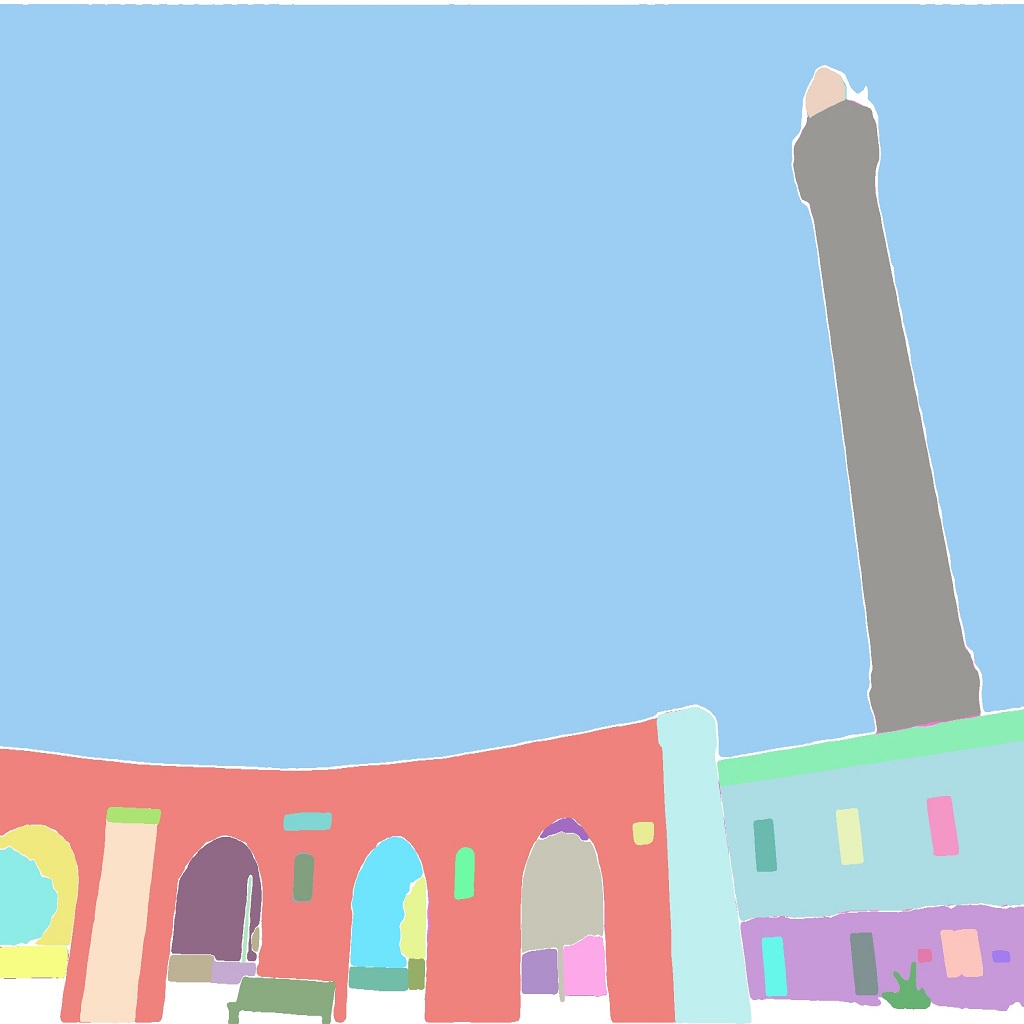}
         \subcaption{Prompt-aware (MobileSAMv2)}
     \end{minipage}

     \caption{Comparison between prompt-free and prompt-aware mask predictions. Prompt-free tends to predict the mask with a non-smooth boundary compared with prompt-aware approaches. For the two prompt-aware approaches, SAM tends to over-segment things while our MobileSAMv2 addresses it due to its object-aware property. Best view in color and zoom in.}
    \label{fig:comparison_qualitative}    

\end{figure*}

As discussed in~\cite{zhang2023faster}, the SegEvery is in essence not a promptable segmentation task and thus can be realized in prompt-free manner. Such an approach has been attempted in~\cite{zhao2023fast} with YOLOv8-seg, which mainly augments YOLOv8-det with a protonet module to generate mask prototype. The intance mask is obtained by convolving the mask prototype with a mask coefficient that has the same length as the prototype dimension (32 by default), which is mathematically a dot product. Here, we point out that the mask decoder of SAM~\cite{kirillov2023segment} also generates the mask by making a dot product between a mask coefficient (called mask token in~\cite{kirillov2023segment}) and a mask prototype (called image embedding in ~\cite{kirillov2023segment}), which have the same (32) dimensions so that the dot product can be computed. Intuitively, the quality of generated mask relies on how well the mask coefficent and mask prototype interact with each other. The mask decoder in~\cite{kirillov2023segment} adopts two-way attention to enable the interaction between the mask prototype and mask coeffcient before performing the final product. Such an interaction is the key foundation for guaranteeing the high-quality mask in SAM. By contrast, there is no explicit interaction between the mask coefficients and mask prototypes in the prompt-free approach. With a single shared mask prototype, it often predicts multiple objects at different regions of the image and thus relies on a bounding box to crop the mask. This can help remove the irrelevant masks outside the box but still fails in yielding high-quality masks as~\cite{kirillov2023segment}, at least partly, due to lack of the interaction between mask coefficient and mask prototype. Even though the prompt-free approach realizes the fastest speed, it results in a non-trivial performance drop (see Table~\ref{tab:comparison_prompt_free}). The less satisfactory performance of the prompt-free approach is mainly attributed to the poor mask boundary (see Figure~\ref{fig:comparison_qualitative}). Compared with prompt-free approach, the two prompt-aware approaches (SAM and MobileSAMv2) generate masks with much more fine-grained boundaries. SAM tends to over-segment things while our MobileSAMv2 alleviates this tendency by utilizing its object-aware property.

\begin{table}[!htp]
\centering
\caption{Influence of the maximum number of prompts on MobileSAMv2 for zero-shot object proposal (mask@1000).}
\label{tab:ablation_max_number_boxes}
\scalebox{0.8}{
\begin{tabular}{ccccccc}
\toprule
max $\#$ of prompts  &all & small & med. & large\\ 
\midrule
384  &59.3 &47.9 &77.1 & 79.9\\ 
\midrule
320  &59.3 &47.9 &77.1 & 79.9\\ 
\midrule
256 &58.5 &46.7 &77.1 & 79.1\\ 
\midrule
192 &56.6 &44.2 &76.0 & 78.8\\ 
\midrule
128 &53.6 &40.2&74.6 & 77.7\\ 
\midrule
64 &44.8 &29.2&68.3 & 75.4\\ 
\bottomrule
\end{tabular}
}
\end{table}

\subsection{Ablation Study} \label{sec:ablation}

With the mask AR@1000 as the metric, we find that our proposed sampling strategy often yields fewer prompts than 1000, which motivates us to explore the influence of the maximum number of (box) prompts in our proposed prompt sampling strategy. The results in Table~\ref{tab:ablation_max_number_boxes} show that increasing the number of box prompts is beneficial for a higher mask AR, however, it saturates after it approaches 320. Therefore, by default, we set the maximum number of prompts in MobileSAMv2 to 320.

\section{Conclusion and Future work}
Orthogonal to the MobileSAM project making SegAny faster by distilling a lightweight image encoder, this project termed MobileSAMv2 makes SegEvery faster by proposing a new prompt sampling strategy in the prompt-guided mask decoder. Replacing the grid-search with our object-aware prompt sampling, we significantly improve the efficiency of SegEvery while achieving overall superior performance. We also demonstrate that our object-aware prompt sampling is compatible with the distilled image encoders in the MobileSAM project. Overall, our work constitutes a step towards a unified framework for efficient SegAny and SegEvery. Future work is needed to seek superior image encoder(s) and object discovery models(s).

{
    \small
    \bibliographystyle{ieeenat_fullname}
    \bibliography{main_mobileSAMv2_arxiv}
}


\end{document}

%% file: preamble.tex
\usepackage[dvipsnames]{xcolor}
